\documentclass[letterpaper,11pt]{article}
\usepackage[utf8]{inputenc}
\usepackage{mathpazo}
\usepackage{setspace}
\usepackage[english]{babel} 

\newcommand{\DJStitle}{The Future of AI is Many, Not One}

\newcommand{\DJSshorttitle}{
\DJStitle
}

\usepackage{amsmath}
\usepackage{amsfonts}
\usepackage{amssymb}
\usepackage[colorlinks=true,linkcolor=black,citecolor=black,bookmarksopen=false, bookmarksnumbered=true,breaklinks=true, pdftex, pdftitle={\DJStitle},
pdfauthor={Daniel J. Singer and Luca Garzino Demo}]{hyperref}
\RequirePackage[round,authoryear,sort]{natbib}
\bibpunct{(}{)}{,}{a}{}{,}

\usepackage{lastpage}
\usepackage{fancyhdr}
\fancypagestyle{plain}{%
\fancyhf{}
\cfoot{\thepage\ of \pageref{LastPage}}
}

\date{February 18, 2026}
\author{Daniel J. Singer\footnote{University of Pennsylvania, Department of Philosophy and Wharton Legal Studies \& Business Ethics}\hspace{\fontdimen2\font} and Luca Garzino Demo\footnote{University of Pennsylvania, Department of Philosophy}}
\title{\DJStitle\thanks{We are extremely appreciative to Patrick Grim and audiences at Hong Kong University, Denison University, and the Philosophy of AI Network for feedback on earlier versions of this work. For feedback or permission to distribute, please contact singerd@phil.upenn.edu and lgarzino@sas.upenn.edu}
}

\makeatletter  
\def\@maketitle{%
  \newpage
  \begin{flushleft}%
  \let \footnote \thanks
    {\large%
    \textbf{\@title} \large \par }
    \vskip .3em%
    {\large
      \lineskip .2em%
        \textrm{\@author}\par
        \lineskip .1em  \textrm{\@date} \par}%
        \end{flushleft}%
 \vskip .1em
    \hrule%
    \null
    }

\renewcommand\section{\@startsection {section}{1}{\z@}%
                                   {-3.5ex \@plus -1ex \@minus -.2ex}%
                                   {2.3ex \@plus.2ex}%
                                   {\normalfont\normalsize\itshape}}
\renewcommand\subsection{\@startsection{subsection}{2}{\z@}%
                                     {-3.25ex\@plus -1ex \@minus -.2ex}%
                                     {1.5ex \@plus .2ex}%
                                     {\normalfont\normalsize\itshape}}
\renewcommand\subsubsection{\@startsection{subsubsection}{3}{\z@}%
                                     {-3.25ex\@plus -1ex \@minus -.2ex}%
                                     {1.5ex \@plus .2ex}%
                                     {\normalfont\normalsize\itshape}}

\renewcommand\paragraph{\@startsection{paragraph}{4}{\z@}%
                                    {3.25ex \@plus1ex \@minus.2ex}%
                                    {-1em}%
                                    {\normalfont\normalsize}}
\renewcommand\subparagraph{\@startsection{subparagraph}{5}{\parindent}%
                                       {3.25ex \@plus1ex \@minus .2ex}%
                                       {-1em}%
                                      {\normalfont\normalsize\itshape}}

\newcommand{\com}[1]{}  
\makeatother   

\let\olditemize=\itemize
\def\itemize{
\olditemize
\setlength{\itemsep}{1pt}3
\setlength{\parskip}{0pt}
\setlength{\parsep}{0pt}
}
\let\oldenumerate=\enumerate
\def\enumerate{
\oldenumerate
\setlength{\itemsep}{1pt}
\setlength{\parskip}{0pt}
\setlength{\parsep}{0pt}
}

\usepackage[protrusion=true,expansion=true]{microtype}
\defcitealias{GrimSingerLandscapes}{Grim and Singer et al. 2013}
\newcommand{\QQQ}[1]{}
\newcommand{\REM}[1]{}

\usepackage{graphicx}
\usepackage{listings}
\usepackage[font=footnotesize,labelfont=bf,skip=2pt]{caption}

\begin{document}

\maketitle

\begin{abstract}\noindent
The way we're thinking about generative AI right now is fundamentally \textit{individual}. We see this not just in how users interact with models but also in how models are built, how they're benchmarked, and how commercial and research strategies using AI are defined. We argue that we should abandon this approach if we're hoping for AI to support groundbreaking innovation and scientific discovery. Drawing on research and formal results in complex systems, organizational behavior, and philosophy of science, we show why we should expect deep intellectual breakthroughs to come from epistemically diverse groups of AI agents working together rather than singular superintelligent agents. Having a diverse team broadens the search for solutions, delays premature consensus, and allows for the pursuit of unconventional approaches. Developing diverse AI teams also addresses AI critics' concerns that current models are constrained by past data and lack the creative insight required for innovation. The upshot, we argue, is that the future of transformative transformer-based AI is fundamentally many, not one.
\end{abstract}

Current transformer-based LLM AI models outperform humans on many tasks, and the ambition of many of those developing and championing AI systems is that they surpass human intellectual ability in all arenas, eventually becoming Artificial General Intelligences (``AGI'') or Artificial Superintelligences (``ASI''). Let's call this ``the ambitious project of AI.'' We're not skeptics about this project, but many are.

Skeptics typically point to three kinds of concerns about the ability of transformer-based LLMs to pull off the ambitious project, including worries about whether these systems can be truly innovative, worries about whether they'll produce monocultures of thought and talk, and worries about whether the opacity of their ``thinking'' processes undermine their ability to explain scientific findings or justify policy interventions. We think there's something right about these objections, but we don't think they foretell the failure of the ambitious project. What they do show, we argue, is that both opponents and proponents of the ambitious project are thinking about the future of AI the wrong way. 

Today, AI is being thought about in a way that's fundamentally ``individual'' or ``singular.'' But decades of research in complex systems, philosophy of science, and organizational behavior have shown that intellectual progress is typically and best made by large, complex, and diverse groups working together, not singular superintelligent researchers or inquirers. In this paper, we'll explain why we should think the same thing applies to the future of AI, why doing so shows that the skeptical worries are misguided, and hence why the future of AI is fundamentally many, rather than one.

In section 1, we draw out the ways in which the future of AI is currently being seen as individual. In section 2, we highlight three kinds of worries AI pessimists have about the ambitious project. In section 3, we spell out why the individual-focused paradigm isn't well-poised to support the ambitious project and explain why using diverse teams of AI agents is a better approach. In section 4, we return to the three worries highlighted in section 2 and show how this new paradigm undermines those worries. In section 5, we show why current multi-agent AI systems don't fully capture what we should be looking for in diverse teams of AI agents and wrap up with some practical takeaways for thinking about the future of diverse teams of AI agents.

\section{We're thinking about AI as one}
AI is being used, thought about, and developed in ways that are fundamentally individual. The field's focus is on singular general AI models, rather than teams or institutions constituted by AI agents. This individual paradigm has shaped the perception and priorities of AI research and embedded itself deeply in the field's practices, and it impacts how new AI models are \textit{developed} and \textit{marketed}, how the models are \textit{measured}, and the \textit{goals} around which development is oriented.

If we look at how AI models are currently being developed, we see research that is primarily organized around a consistent ambition to build a single, maximally powerful model, rather than a team, hierarchy, or community of models that work together. It feels like every day now that an established player or a new startup announces a new or updated singular foundational model, a large, general-purpose system trained on vast datasets and designed to be flexibly adapted to various tasks \citep{bommasani2021opportunities, schneider2024foundation, liu2024position, schrepel2024competition, pyzer2025foundation}. Prominent examples include GPT-5.4 (March 2026), Grok-4.20 (March 2026), Claude Opus 4.6 (February 2026), Gemini 3 Pro (November 2025), Qwen 3.5 (February 2026), Llama 4 (April 2025), and DeepSeek V3.2 (December 2025). Singular, massively capable models like these have become the  dominant product line of the industry, and the public face of AI itself as millions of AI users interact daily with individual models \citep{maslej2025artificial, Maslej2024AIIndex, liu2024earth}.

The approach of focusing on individual foundation models isn't an accident. It gets  theoretical justification from scaling laws, which show that model performance improves predictably as functions of parameter count, dataset size, and computational resources \citep{kaplan2020scaling, hoffmann2022training, muennighoff2023scaling}. These laws provide an empirical recipe for progress: more scale = better model. And while we don't doubt the success of this approach, we do want to highlight that the focus on scaling laws has led to a narrow, singular way of thinking about how to make better AI, where the prevailing answer has been to scale up a single model, rather than e.g., to move to using a collection of models.

The individualistic orientation toward our future with AI can also be seen in how AI progress is measured. To measure and compare progress, AI developers, customers, and outside experts all use AI benchmarks. These benchmarks are fundamentally standardized tests on which models are scored and ranked. For example, the Massive Multitask Language Understanding benchmark evaluates models across 57 subjects, from elementary mathematics to professional law \citep{hendrycks2020measuring}, the Holistic Evaluation of Language Models provides more comprehensive assessment across dimensions of accuracy, fairness, and robustness \citep{liang2022holistic}, and several popular websites like  LiveBench \citep{livebench}, LMSYS Chatbot Arena \citep{chiang2024chatbot}, and Artificial Analysis \citep{ott2022mapping} all provide up-to-date evaluations on different tasks. Although these benchmarks take very different approaches to measuring the performance of LLMs, what they share in common is that they measure the performance of \textit{single} models. 

Because benchmarks are used to measure progress, they act as a kind of developmental infrastructure that constrains innovation. In AI development, topping a leaderboard is the clearest and most public signal of technical superiority. It is a press event for a company, a career-defining milestone for a research team, and a potent validation of particular design choices. It also directly translates into economic value by attracting investment, justifying vast capital expenditures, exciting top-tier talent, and drawing in new customers. The result is that benchmarks of singular models produce a powerful incentive structure aimed at a singular goal: producing the next individual model that can climb to the top of the charts.

The focus on singular models as the future can also be read directly off the hopes, aspirations, and even planning documents of major players in the field. For most, the conceptual horizon of AI research is dominated by the image of a single, superintelligent agent, typically labeled ``AGI'' or ``ASI,'' that represents the ultimate culmination of AI development. OpenAI, for one, makes this very explicit, titling their planning document ``Planning for AGI and beyond'' \citep{openai2023planning, bubeck2023sparks}. Corporate strategies, research agendas, and funding decisions are increasingly organized around the race to achieve this general/superintelligence milestone. Even forecasting firms construct their scenarios around the timeline to \textit{the} first AGI, as though AI history will culminate in the arrival of one decisive actor \citep{roser2023ai, lifland2023ai}. The terminology used here reveals the underlying assumption that there will be a singular model that constitutes "the" AGI.

Moreover, the field's most prominent safety and governance concerns are structured entirely around managing a singular AI entity. The majority of AI alignment research is framed in terms of ensuring that one superintelligent system pursues human-compatible goals \citep{soares2014aligning, ji2023ai}. This is predominantly modeled as an interaction between humans and a single other player---the potentially misaligned AI \citep{amodei2016concrete, christiano2017deep, christian2020alignment, hendrycks2020aligning}. Consequently, even the most ambitious alignment proposals take as their starting point the assumption of one central agent whose misaligned actions could prove catastrophic \citep{ouyang2022training, ngo2022alignment, ji2023ai}.

What this shows is that the paradigm for the future of AI is fundamentally centered on individual AI models. Methods of production are designed to build individual massive models, benchmarks are designed to rank individual models, and long-term aspirations are framed around the arrival of a superintelligence instantiated in an individual model. This drives innovation toward a singular goal, producing the next top individual model---an achievement that translates directly into press coverage, investment, talent recruitment, and customer acquisition --- and it provides a developmental infrastructure that includes powerful economic and reputational costs for exploring alternative approaches. 

\section{Three worries for AI general/superintelligence}
We're optimistic about the ambitious project, but not everyone is. There are three main worries that more pessimistic members of our community have about the ability of current AI models to reach the AGI/ASI benchmark. 

The first worry is about innovative or transformational thinking. The concern is that transformer-based LLMs, because they're constrained by their training data, necessarily lack the capacity for the creative and visionary insights that have characterized many of the most important human intellectual pursuits of the past \citep{ding2025generative, Felin2024-yj, kapoor2024reforms, runco2023ai}. \cite{Felin2024-yj}, for example, argue that a transformer-based AI in the late 19th century that had access to the overwhelming evidence, scientific consensus, and repeated failures of powered flight, would have concluded that powered flight was impossible (agreeing with the \textit{New York Times}, which had recently concluded that the project would take ``one million to ten million years''). But as we know, the (human) Wright Brothers succeeded in that task despite the overwhelming evidence. According to \citet{Felin2024-yj}, this creative, world-changing way of thinking is inaccessible to transformer-based AIs. As backward-looking prediction machines, LLMs can only repeat patterns in their training data, leaving them incapable of replicating human creative and visionary leadership, these authors argue.

The second worry is that AGI/ASI will create what \cite{messeri2024artificial} call ``scientific monocultures.'' The concern is that once AI systems become more objective and reliable than human judgment (or at least appear to be so), the public (including scientific researchers) might be inclined to privilege the questions and methods most legible to AI, mistaking the tractable for the true. The result is an illusion of exploratory breadth and objectivity: the public and scientists believe they are examining all plausible hypotheses when, in fact, they are confined to the subset of problems that align with the model's strengths. So, the worry goes, by integrating very intelligent models into public discourse or research, we risk reducing the marketplace of ideas to a homogenized version of public discourse and research, narrowing the range of ideas discussed and making the intellectual community (including science) more vulnerable to collective error.  

The third worry is about explanation and reasons opacity. The concern is that current transformer-based AI systems can't produce real explanations or reasons for what they believe (or say they believe). Critics argue that because deep learning models function as black boxes, the parameters and non-linear correlations that drive their outputs are fundamentally inaccessible to human understanding, creating a problem of epistemic opacity \citep{Creel2020-CRETIC, Rudin2019-zz, Zerilli2022-ZEREML, Humphreys2009-HUMTPN, Lipton2018}. Unlike human researchers, who can ideally provide the reasoning that led to a conclusion, AI models neither arrive at conclusions by a traditional process of reasoning nor can they offer up the probabilistic sequence of tokens that lead to their conclusions. Any so-called ``explanation'' that's proffered by a transformer-based LLM is, at best, a plausible-sounding post-hoc rationalization.\footnote{\citet{Turpin2023} demonstrate this disconnect empirically, showing that ``Chain of Thought'' explanations are frequently unfaithful to the model's actual process; for instance, if a model is biased toward an incorrect answer by its input, it will often hallucinate a fictitious logical explanation to justify that biased conclusion.} Because scientific and public discourse both rely on the giving and assessing of explanations and justifications \citep{Sullivan2022-SULUFM, Zednik2019-ZEDSTB}, the worry goes, even if AI systems become incredibly reliable predictors, they won't be able to fully participate as reason-givers in our shared justificatory practices. At best, they'll be able to supply opaque outputs around which humans must still do the work of articulating, evaluating, and revising the actual explanations that guide theory choice and policy.

As we'll argue below, if we abandon the individual paradigm and instead think of AI systems as consisting of diverse teams of reasoners, we can answer each of these worries while gaining additional benefits associated with having groups, rather than individual, inquirers.

\section{Why the future of AI is many}

Decades of research in complex systems  \citep{hong2001problem, Hong16112004, Page2008-xf, Singer2019-SINDNR}, philosophy of science (\citetalias{GrimSingerLandscapes}, \citealp{Weisberg2009-WEIELA}, \citealp{Zollman2010-ZOLTEB-2}), organizational behavior \citep{lazer2007network, phelps2012knowledge, grim2024epistemic}, and computational social science \citep{becker2017network, centola2023experimental} have shown that intellectual progress is typically made by intellectually diverse groups working together, not singular brilliant researchers working on their own. In this section, we'll argue that the same lesson applies to artificial intelligences. The benefits of groups are at least as strong for AI agents as they are for human researchers, and AI agents bear fewer costs of working in groups. In the literature, there are many purported benefits of epistemically diverse groups, but here we'll focus on three. Those are that they (\textit{i}) explore a broader region of the problem space; (\textit{ii}) avoid premature convergence on misleading solutions; and (\textit{iii}) divide the work of being accurate and being innovative. 

The history of science is replete with cases where the most successful periods are marked by competing methods and parallel lines of inquiry rather than the dominance of a single approach. Consider four massive breakthroughs, each of which reshaped their fields: the development of the COVID-19 vaccines \citep{kariko2024breaking, szabo2022covid, saleh2021vaccine}, the rival programs that uncovered CRISPR \citep{lander2016heroes, ledford2016unsung}, the competing models that revealed DNA’s structure \citep{scott2012discovery, schindler2008model}, and the long contest among paradigms in AI itself \citep{fisher2025making, dhar2024paradigm}. What these cases share is not a moment of singular genius, but a pattern of distributed exploration: many paths were explored at once, with different researchers seeing and exploring possibilities that others missed or doubted. The winning strategy in each case wasn't obviously correct at the outset. It succeeded because the scientific community, as a whole, spread its bets widely enough to find it.

This reveals the first benefit of epistemic diversity: hypothesis breadth. Diverse groups explore more of the problem space than any individual could. Philosophers of science call this organization the \textit{division of cognitive labor} \citep{Kitcher1990-KITTDO, psk:advancement, Weisberg2009-WEIELA}.
Division of cognitive labor allows communities to cover ground no individual could. But there is also a deeper, mathematically grounded result about how diversity operates \textit{within} groups. \cite{Hong16112004} prove that cognitively diverse groups, when confronted with the same sufficiently difficult problem, can be expected to outperform expert groups (and single experts) because diverse groups generate and consider a wider range of solutions \citep{Singer2019-SINDNR}. Diverse problem solvers approach challenges with different tools: their non-overlapping heuristics make them move differently through the problem space. Where one researcher's method stalls, another's makes progress.

Crucially, in these cases, diversity enables genuine collaboration through complementary cognitive styles. In cognitively diverse groups, one agent's way of approaching a problem can make progress that another agent, reasoning differently, might not have been able to make themselves but is uniquely able to extend \citep{Hong16112004}. For instance, the discovery of the structure of DNA was made possible by Rosalind Franklin’s X-ray diffraction data. This evidence revealed structural constraints that Watson and Crick, working with a different set of theoretical tools, were then able to develop into a full double-helix model \citep{schindler2008model, scott2012discovery, pray2008discovery}. In such cases, individuals' complementary strengths allow the team to reach solutions that none of the members could have found alone. Agent-based models in philosophy of science and computational social science highlight the same dynamic: diverse strategies reduce blind spots and create opportunities for sequential improvement across members of the community \citep{Weisberg2009-WEIELA, GrimSingerLandscapes, Singer2019-SINDNR, becker2017network, centola2023experimental}. In this way, diversity creates productive synergies where one line of inquiry generates precisely the insight another needs to reach a breakthrough.

The long dispute over the cause of peptic ulcers illustrates the second benefit of epistemic diversity: sustained exploration. For decades, two hypotheses competed: one centered on excess stomach acid, the other on bacterial infection \citep{schwarz1910ueber, Thagard1999-THAHSE}. Early studies appeared to support the acid hypothesis, and a single influential mid-century study seemed to rule out bacteria altogether \citep{palmer1954investigation}. These early signals were strong enough to pull much of the community into a settled view, and research on the bacterial hypothesis nearly disappeared. Yet small pockets of dissent kept the alternative alive. And this line of inquiry eventually produced decisive evidence for bacterial infection, the correct answer \citep{marshall1984unidentified}. Had the community been more uniform, the bacterial hypothesis might have disappeared entirely, leaving millions suffering from a treatable disease.  

What this case shows is that intellectual communities face a serious risk of premature convergence. In homogeneous groups, early promising results can lock inquiry onto a single approach even when that approach is ultimately inferior. Work on networked communities demonstrates that tightly connected, methodologically uniform groups amplify their earliest signals, locking themselves into suboptimal paths before competing hypotheses have had a fair chance to develop \citep{Zollman2007-ZOLTCS, Zollman2010-ZOLTEB-2, lazer2007network}. This dynamic mirrors groupthink, where social and methodological homogeneity accelerates consensus at the cost of error correction \citep{sunstein2014making, tetlock1992assessing}.

By contrast, diverse groups of researchers are more persistent than individual experts and teams of experts---especially when structured through semi-independent clusters. Because their members approach the same evidence with different assumptions, methods, and inferential styles, they keep multiple hypotheses alive long enough for their strengths and weaknesses to become visible \citep{Zollman2007-ZOLTCS, Zollman2010-ZOLTEB-2, lazer2007network}.

The third benefit addresses a fundamental tension in inquiry: the conflict between accuracy and innovation. Groups are often successful because some members track evidence carefully and update conservatively---these are the kind of scholars who ensure knowledge remains reliable and well-grounded. But communities thrive also by having members who generate speculative hypotheses, explore unfamiliar models, and push beyond the most likely paths---those are the kind of thinkers who drive genuine breakthroughs. The problem is that these aims pull in opposite directions. Efforts to be maximally accurate tend to keep inquiry close to established results, while efforts to be genuinely creative push toward more uncertain territory \citep{zollman2018credit, vsevselja2023agent, o2023modelling, stadler2014solutions, toyokawa2014human}. And it's exceptionally difficult for any single investigator to embody both tendencies at once. Someone who prioritizes reliability will miss the bold leaps that lead to discovery. Someone who goes for creative leaps will often be wrong. 

Communities resolve this tension not by forcing individuals to balance these competing demands, but by distributing the demands across different investigators. In particular, the lesson from science is that the optimal community is one where some researchers are more accuracy-focused, and still others are more creative and outside-the-box thinkers. A diverse group can combine these tendencies without forcing any single agent to balance them internally. We might call this benefit ``parallel pursuit of competing aims,'' as this structure allows a community to be both reliable and inventive. It preserves the accuracy of its most conservative members while drawing on the boldness of its most exploratory ones.

These three benefits---hypothesis breadth, sustained exploration, and parallel pursuit of competing aims---explain why diverse communities routinely outperform even the most capable individuals on complex problems. The same logic applies directly to AI. Nothing in these mechanisms depends on specifically human features. The force of these results comes from the mathematical structure of difficult problems themselves. Complex problems are filled with solutions that look promising but lead nowhere. Due to this structure, there is just no reliable way to know in advance what works and what doesn't. Even advanced AI models, armed with sophisticated reasoning abilities, will get stuck on these misleading paths. But what traps one model might not trap another built or trained differently. This is why differently structured models, working together, can discover solutions that any single model would miss.

Single investigators (and single models) will always face a trade-off between accuracy and innovation. The features that make a researcher dependable also constrain its ability to pursue uncertain leads. And single researchers will always search the problem space using their particular architecture and training, converging on whatever solution those constraints make visible. A team of diverse researchers can explore more of the landscape, avoid settling too quickly on attractive but misleading ideas, and divide epistemic labor in ways no individual  can. Designing AI as a community of diverse models makes it possible to realize these structural advantages: a system that can be cautious and bold at the same time.

Moreover, AI communities can capture these benefits while avoiding the costs that make diversity challenging for human researchers. Diverse groups of humans pay a real price for the sustained disagreement that epistemic diversity requires: coordination overhead, communication friction, and the interpersonal tensions that arise when researchers with different methods and assumptions must work together \citep{minbaeva2021beyond, zhan2015re, carter2017double}. AI communities don't directly inherit these burdens. Differences in architecture or training do not produce interpersonal friction for AI models. Parallelization makes it possible for AI agents to explore alternative approaches at scale: what would be a years-long delay in human research becomes cheap in computational terms. As a result, AI communities can preserve all three benefits of diversity while replacing the usual human costs with manageable, often negligible, computational costs.

For these reasons, we think it's time for a paradigm shift that moves us from the singular pursuit of an individual superintelligent model to the design of diverse AI communities. 

\section{How the ``many'' approach resolves the skeptics' worries}

Reframing how we think of the future of AI by shifting from the individual paradigm to a more group- or community-based one also undermines the most prominent objections to AI-driven innovation. Those concerns, it turns out, are concerns about the limits of \textit{individual} AI systems, ones that can be easily overcome by structuring AI systems as communities of models.

First, the innovation worry: Skeptics argue that transformer-based LLMs, constrained by training data, cannot produce the creative, paradigm-breaking insights that characterize transformative human achievement \citep{Felin2024-yj, ding2025generative, kapoor2024reforms, runco2023ai}. An AI model trained on the overwhelming late-19th-century evidence that powered flight was impossible would have concluded exactly what most experts concluded: that the project would take millions of years, if it could be done at all.

But this worry incorrectly assumes that innovation is something that springs from individuals, rather than communities. The Wright Brothers weren't working in isolation. They were embedded in a broader scientific and engineering community, most of whom were pursuing very different approaches. The Wrights succeeded not because the entire community abandoned conservative, evidence-based reasoning, but because the community's structure allowed them to pursue a high-risk, unconventional approach while others maintained more cautious lines of inquiry. What made innovation possible was the division of cognitive labor: most researchers tracked the evidence conservatively, while a few pursued long-shot alternatives.

To encourage innovation, AI communities should replicate this structure. We can train and instruct some models to take creative leaps and be contrarian. But we shouldn't want every model to be a risk-taking contrarian since we need the group to also care about accuracy. Groups with both kinds of models should then be structured to reproduce the institutional design that made the Wright Brothers' breakthrough possible. Properly designed, the group would allow the successful ideas of the more creative models to filter through while filtering out less productive ideas. AI need not choose between accuracy and creativity, since properly structured communities of AI eliminate this dichotomy.

The second worry was the monoculture concern: once AI systems become authoritative sources of knowledge, we risk privileging the questions and methods most legible to those systems, narrowing the space of inquiry and creating intellectual homogenization \citep{messeri2024artificial}. The concern points to a real phenomenon: groupthink, the tendency for homogeneous groups to converge prematurely on shared assumptions and accept expected conclusions too swiftly, amplifying biases and suppressing alternative approaches \citep{sunstein2014making, tetlock1992assessing}. This is what happened in the 2008 financial crisis: nearly every major economic model failed to predict it, not because economists were unintelligent, but because they shared the same assumptions about market stability and risk distribution \citep{besley2009letter, colander2009financial, kirman2010economic}. The models were all looking in the same direction, seeing the same patterns, missing the same warning signs. 

But groupthink is a problem of certain group structures, not of intelligence per se. And it's a problem we already know how to solve. Scientific communities have faced this risk throughout history and developed institutional mechanisms to counteract it. They sustain epistemic diversity by rewarding successful contrarians with outsized credit, ensuring that disagreement remains intellectually viable and professionally profitable \citep{volmar2025mavericks, heesen2019credit, heesen2017communism, rubin2021priority, Strevens2003-STRTRO-5}. This reward system in science ensures the division of cognitive labor, hypothesis breadth, and sustained exploration \citep{copeland2019serendipity, Mayo-Wilson2011-MAYTIT, Strevens2003-STRTRO-5}.

The same structural principles should be used to prevent AI monocultures. A community of AIs should be organized as a competitive ecosystem that rewards divergence. Teams of models would not be optimized toward a single metric of success, but toward varied and partly conflicting objectives whose interaction produces discovery. These incentives would drive diversity in practice: to maximize their expected success, models would be driven to adopt distinct heuristics, pursue uncorrelated hypotheses, and search different regions of the conceptual landscape. Then, we'd avoid monocultures by structuring incentives that make diversity sustainable. 

Moreover, we can build genuine cognitive diversity directly into the system, as we'll discuss in the next section. Models can be trained with different architectures, different loss functions, different training datasets, and different reasoning priorities. AI models can be designed from the ground up to reason in fundamentally different ways. Where singular AI narrows the space of inquiry and risks groupthink, diverse AI communities can expand it by design. Homogenization is not an inherent risk of AI. It's a consequence of singular design choices that we can avoid. So, again here, once we switch to a group-centered approach to AI, the worry disappears.

The third worry was about opacity: Skeptics claim that AI systems cannot provide genuine explanations or reasons for their outputs, which undermines their ability to participate in the justificatory practices essential to science and policy \citep{Sullivan2022-SULUFM, Zednik2019-ZEDSTB, Creel2020-CRETIC}. The thought is that because transformer-based models arrive at their conclusions via a process that cannot be reported back to us, they cannot offer anything from their reasoning as an explanation or justification of their conclusions \citep{Humphreys2009-HUMTPN, Lipton2018, chesterman2021through}. All we get instead is plausible-sounding post-hoc rationalizations that may bear little relation to the actual computational process that produced the result \citep{Turpin2023, Lipton2018, Zerilli2022-ZEREML}. For scientific discourse and policy deliberation, this seems like a fatal limitation, as both enterprises depend on the giving and assessing of reasons.

But this concern misunderstands what explanations and reasons actually do in scientific and policy contexts. Explanations and reasons don't (and shouldn't) purport to be faithful reports of the cognitive processes individual scientists actually used. Even human scientists can't reliably introspect their own reasoning. Explanations and reasons instead serve a \textit{social} function: they allow the community of researchers to evaluate claims, identify points of disagreement, propose alternatives, and build on one another's work. What matters isn't whether an explanation captures a private thought process, but whether it provides colleagues with the resources to assess conclusions, understand implications, and determine where to push back or extend the finding.

AI teams enable the production of these kinds of explanations. In a properly structured AI community, multiple models approaching the same problem with different architectures and training will make their disagreements visible and interpretable. Like in our current scientific institutions, scientists defending ideas must produce not only the ideas but the evidence and reasoning for them as well, and that evidence and reasoning is broadly considered by the community before the ideas are taken up. Similarly, in AI groups properly structured, the interaction between different models---where one flags uncertainties another glossed over, or where competing approaches surface different implicit assumptions---will produce the kind of explanation that matters for scientific discourse and policy decisions. These explanations come from communities structured to support contestation, uptake, and revision, not from individual researchers faithfully reporting the causal process that gave rise to their conclusions.

Taken together, these responses show that the most prominent objections to the ambitious project of AI stand only if we think in individualistic terms. If we want the ambitious project of AI to succeed, we need to organize teams of models that think differently and work together.

\section{What diverse groups of AI agents look like}
At this point, we hope to have convinced you that the ambitious project of AI requires a shift from the individual to the many. But how do we actually build the ``many''? 

You might think that the industry is already moving toward systems that are ``many.'' Most of the current state-of-the-art LLMs use a ``Mixture of Experts'' (MoE) architecture, wherein a routing mechanism directs different queries to different ``experts'' contained within the model. But MoE models like these aren't the kind of ``many'' that we're looking for. In these models, the ``experts'' are trained simultaneously under a single loss function, share a common backbone or embedding space, and update their weights to minimize a global error metric. As a result, the ``experts'' share the same inductive biases and blind spots. They do not offer the uncorrelated error modes that are one of the central elements needed for epistemic diversity to yield benefits \citep{Hong16112004, Page2008-xf}. Moreover, these ``experts'' don't work together in any real sense. Even though MoE models do allow single tokens to be processed by multiple experts, which results in an output that is genuinely a mixture, MoE models are best seen as dividing up tasks and doing the work in parallel, not together. In other words, they represent a sparse activation of a single probability distribution, not a collaboration of distinct reasoners.

Similarly, while it's a step in the right direction, we shouldn't think that it's good enough for a singular, massive model to simulate a diverse team simply by being prompted to ``play devil's advocate'' or ``critique its own reasoning.'' While these are useful prompting techniques, all of a model's internal critics will still be bounded by the model's own priors and training data. When a model critiques itself, it performs inference over the same probability distribution that generated the original output (and its initial error). If a model has a systemic gap in its training data --- say, regarding a specific chemical interaction --- its internal critics will generally share that gap. So doing this doesn't provide a robustly independent check; it only offers the most likely critique that a model of its type would generate.

To achieve the benefits of diverse groups of inquirers outlined in Section 3, we should focus on two big facets of artificial epistemic communities: (1) diversity in the constitution of the agents and (2) variation in how the interaction between those agents is structured. 

\subsection{Three Dimensions of Epistemically Diverse AI Agents}
In thinking about epistemic diversity for AI agents, we should focus not just on having agents that produce different outputs. We should focus on doing that in ways that are both likely to contribute to better overall group outcomes while also being computationally tractable and feasible. We identify three dimensions of variation that we should treat as most salient in doing that project. 

\textbf{Dimension 1: Stochasticity.} Generative AI models are fundamentally probabilistic. Before giving an output, they create a probability distribution over several options for the next tokens. The simplest way to make diverse agents leverages this stochasticity, creating teams of agents that are identical in training and architecture but divergent in output. By manipulating inference-time hyperparameters --- specifically Temperature (which flattens the probability distribution to allow lower-likelihood tokens), Top-k (which limits the sample pool), and Top-p (nucleus sampling, which dynamically truncates the tail of the distribution) --- we can force multiple instances of the same model to explore options it might not otherwise see. In the language of complex systems, this functions as ``local search.'' If one agent gets stuck in a local optimum (a repetitive loop or a common misconception), a high-temperature counterpart might ``jump'' out of that valley simply by sampling a less probable token, which means the group as a whole can do better.

\textbf{Dimension 2: Perspective.} The second dimension leverages the ``In-Context Learning'' capabilities of modern LLMs. Through the use of prompt and persona engineering, we can steer a model's attention mechanism, forcing it to weigh features of an input differently. This allows us to instantiate agents that, while sharing an underlying LLM ``brain,'' adopt radically different cognitive stances. E.g., consider a team analyzing a complex social crisis. If we prompt one agent to reason as a neoclassical economist, its self-attention mechanism will prioritize tokens related to incentives, efficiency, and market equilibrium. If we prompt another agent to reason as a comparative political theorist, it will attend to tokens regarding power structures, historical precedence, and cultural narratives. In Section 3, we noted that the benefits of epistemic diversity come, not just by adding randomness, but by adding heuristics and the ability of distinct researchers to work together and build on one another's ideas. AI communities with perspective diversity, in this sense, realize these mechanisms (and their related benefits) without having to use more than one underlying model. 

\textbf{Dimension 3: Constitution.} The final dimension of diversity for AI agents involves varying the underlying model. This occurs at two levels: data and architecture. Models are bounded by their training corpora. An agent pre-trained primarily on legal texts will have a fundamentally different map of the conceptual relationships between words than an agent trained on biological literature. So, AI agents can vary in their constitution by being trained on different data. Agents can also vary in their underlying architectures. Different neural architectures impose different inductive biases. Transformer models (which generally excel at long-range pattern matching) will make different kinds of reasoning errors than a State Space Model (like Mamba) or a Neurosymbolic system (which integrates neural networks with symbolic logic). Agents that vary in this way vary in a very deep sense. These agents will process information with fundamentally different mathematical techniques and their error modes are far less likely to be correlated. E.g., where a Transformer hallucinates due to a probabilistic association, a Neurosymbolic agent is unlikely to be susceptible to the same failure mode. By building teams that vary across this dimension, we create a system where the weaknesses of some agents can hopefully be covered by the strengths of another, again leaving the team as a whole better off.

\subsection{Institutions of Interaction: From Teams to Ecosystems}
Just having diverse AI agents isn't enough. To reap the benefits of epistemically diverse groups, the agents must also work together well. Just as human group inquiry comes in many forms, including everything from tight-knit labs to global scientific institutions, AI collectives can be instantiated along a spectrum of social cohesion.

At one end of the spectrum are \textbf{flat teams}. These are small, non-hierarchical groups of agents working on a shared objective, mimicking a jury or a chairless committee. Here, the primary goal of interaction is accuracy and error correction. So, the interaction protocols might include ensemble voting to filter out hallucinations (leveraging the Law of Large Numbers) or adversarial debate, where agents are prompted to find flaws in each other's proposals and then help rebuild them. In groups that work like this, it's easy to see how using groups of AI agents addresses the AI skeptic's opacity worry: the explanation for an output or decision is not a trace of weights, but the transcript of the debate between the agents.

Further along the spectrum are \textbf{hierarchies}. For problems that are too complex to be solved in a single inference pass, such as designing a novel protein or writing a software suite, agents can be arranged in hierarchies mimicking a research lab or project team with a manager. Here, the aim of the structure is not just error-checking, but decomposition and synthesis. So, in a setup like this, a ``Principal Investigator'' agent --- perhaps a large, general-purpose model with a large context window --- acts as the architect. It breaks a high-level goal into sub-components and assigns them to specialized ``post-doc'' or ``grad student'' agents (e.g., a Coder agent, a Literature Review agent, a Data Analyst agent). The hierarchy creates a mechanism for coherence. While the specialized agents focus on their narrow tasks, the PI agent integrates their outputs, resolving conflicts and ensuring the parts fit into the whole. This allows the system to tackle long-horizon tasks that usually cause singular models to drift off-topic.

Finally, at the other end of the spectrum, are \textbf{scientific ecosystems}, which are functionally communities that might have distributed or even competing goals. In this kind of structure, distinct groups of agents do not always communicate directly. One group might publish a ``(artificial) paper" (a finalized output) to a shared database, where an independent group of agents, blind to the first group's process, reviews or attempts to replicate. ``(Artificial) granting agencies'' might accelerate the research and increase the visibility of some subgroups while ignoring others. And ``(artificial) research institutions'' might support the research of some scholars and allow them to replicate (via ``AI grad students and post-docs'').

Broader ecosystem structures like these can realize the parallel pursuit of competing aims discussed in Section 3. By structurally isolating different groups, we allow for a division of labor at the community level: some clusters of agents (using high temperature and novel prompting) can pursue wild, low-probability hypotheses, while others (maybe using formal logic architectures) act as conservative verifiers. This kind of structure is also best poised to produce innovation. If these agents were all in a single hierarchy, a risk-averse ``manager'' agent might suppress the wild hypothesis before it could be developed. But in an ecosystem, the risk-takers have the space to succeed against the odds (or, more often, just fail but in ways that don't take down the entire ecosystem). This mirrors our take on the Wright Brothers example from Section 4: the ecosystem structure allows for radical risk-taking in one corner of the graph without sacrificing the community's overall grounding in established reality.

\bigskip

By focusing on varying individual agents and the institutions they interact in, the engineering goal shifts from minimizing a single loss function to maximizing the collective intelligence of a group. Just as the institution of science survived the errors of individual scientists to produce the Wright Brothers’ breakthrough, an ecosystem of diverse AI agents can transcend the hallucinations, blind spots, and opacity of the singular models of which it is composed.

\noindent\rule{4cm}{0.4pt}

\medskip

The ambitious project of AI is at a developmental crossroads. If we continue in the individual paradigm, chasing a singular, omniscient AGI, we risk hitting the limits of monoculture, stagnation, and opacity that AI skeptics fear. But if we embrace the lessons of complex systems and the history and philosophy of science, a more optimistic path emerges. By accepting that innovation and increases in intelligence are fundamentally driven by groups, rather than individuals, we can engineer systems that divide cognitive labor, hedge against error through diversity, and sustain the tension between accuracy and innovation. The critics are right that a single transformer cannot replicate the full sweep of human discovery. But they are wrong to think that AI cannot. The ambitious project of transformative AI is reachable, but only if we see it as fundamentally many, not one.

\end{document}